# Privacy-Preserving Chaotic Extreme Learning Machine with Fully Homomorphic Encryption

Syed Imtiaz Ahamed[1,2] and Vadlamani Ravi[*]

[1]*Centre for AI and ML,*

*Institute for Development and Research in Banking Technology Castle Hills,*

*Masab Tank, Hyderabad 500057, India*

[2]*School of Computer and Information Sciences (SCIS), University of Hyderabad,*

*Hyderabad 500046, India*

*syedahamed@idrbt.ac.in; vravi@idrbt.ac.in*

## Abstract

The Machine Learning and Deep Learning Models require a lot of data for the training process, and in some scenarios, there might be some sensitive data, such as customer information involved, which the organizations might be hesitant to outsource for model building. Some of the privacy-preserving techniques such as Differential Privacy, Homomorphic Encryption, and Secure Multi-Party Computation can be integrated with different Machine Learning and Deep Learning algorithms to provide security to the data as well as the model. In this paper, we propose a Chaotic Extreme Learning Machine and its encrypted form using Fully Homomorphic Encryption where the weights and biases are generated using a logistic map instead of uniform distribution. Our proposed method has performed either better or similar to the Traditional Extreme Learning Machine on most of the datasets.

## 1. Introduction

In every field such as healthcare, finance, education, and various other fields the organizations collect and store a lot of data in databases which involves private information and is freely utilized to build better ML models. In such scenarios, privacy and security become the major concerns and they cannot be simply ignored. Organizations need to be careful about the protection of customers' Personal Identifiable Information (PII) and also try to find a solution that would help them to analyze the data at the same time.

---

[*] Corresponding author.

General Data Protection Regulation (GDPR) in [1], which is one of the toughest privacy and security laws was brought into action by the European Union (EU) on May 25, 2018. According to the law, the organization would be fined almost millions of euros if they violated the privacy and security standards. Two more privacy laws namely California Consumer Privacy Act (CCPA) and Personal Data Protection Act (PDPA) were enacted in California and Singapore respectively. The CCPA law provides the consumers in California with the right to know about every detail that a business collects from its clients, to delete the collected information, and also to opt-out from their data being sold out as explained by Stallings in [2]. The enactment of PDPA helped in the protection of personal data by Chik in [3].

The implementation of such strict laws resulted in the utmost security and privacy to the customer data and identity but became a problem for the organizations as they could no longer use the private data easily. This problem can be solved by using the different methods provided by PPML and building Machine Learning and Deep Learning models which can benefit the organization and at the same time guarantee that there will be no compromise with the customer's data and identity.

One of the techniques of PPML is Secure Multi-Party Computation in which multiple data owners can collaboratively train a model without actually knowing anything about each other's input data and will only be able to access their respective outputs. The drawback of this technique is the high computation or communication overhead as mentioned in [4]. The other technique is Differential Privacy which allows working on the personal information of the people without disclosing their identity but DP might result in a loss in the model accuracy.

Homomorphic Encryption is one more approach to secure the data which allows performing the computations on the encrypted data without decrypting it. They are three types of Homomorphic Encryption namely: Partial Homomorphic Encryption (PHE), Somewhat Homomorphic Encryption (SWHE), and Fully Homomorphic Encryption (FHE). PHE allows an unlimited number of similar types of operations (either addition or multiplication), SHE allows a limited number of both operations, and FHE allows an unlimited number of arithmetic operations on the encrypted data.

Generally, FHE is considered to be better than other techniques in terms of security but it is computationally very expensive. In this paper, we propose a Chaotic Extreme Learning Machine and FHE-based privacy-preserving Chaotic Extreme Learning Machine (ELM) for the classification task. Here the meaning of Chaotic means that the parameters like weights and bias are generated using the logistic Map. Thus we designed and implemented the secure Chaotic ELM by ensuring that the data and all the parameters in the network are fully homomorphic encrypted and also we get the results in an encrypted format.

The remaining part of the paper is structured as follows: In section 2, related work regarding homomorphic encryption in Machine Learning is discussed. The proposed methodology is explained in Section 3 and the description of the datasets is presented in section 4. The results are discussed in Section 5 and finally, the conclusions are presented in Section 6. Appendix A consists of Tables presenting the features of datasets.

## 2 Literature Survey

The concept of privacy-preserving the data has resulted in the application of privacy preservation in most Machine Learning Algorithms. To begin with, Nikolaenko, et.al., [5] proposed a privacy-preserving ridge regression by combining Yao garbled circuits with linear homomorphic encryption. Later, Chabanne, et al., [6] proposed a fully homomorphic encrypted Convolution Neural Network combined with the solution of Cryptonets [7] along with the batch normalization principles.

Chen, et al., [8] implemented fully homomorphic encryption Logistic Regression using the Fan-Vercauteren scheme implementation in the SEAL Library. In a homomorphically encrypted logistic regression, an ensemble gradient descent method to optimize the coefficients was introduced by Cheon, et al., [9] which resulted in the reduction of time complexity of the algorithm.

Qiu, et al., [10] proposed a Privacy-preserving Linear Regression model consisting of the Paillier Homomorphic Encryption and data masking technique which includes multiple clients and two non-colluding servers. Bonte & Vercauteren [11] implemented Privacy-Preserving Logistic Regression where somewhat homomorphic encryption based on the scheme of Fan & Vercauteren [12] was used.

Bellafqira, et al., [13] implemented a secure Multi-layer perceptron using the Paillier cryptosystem and homomorphically encrypted data is trained on the cloud. Later, Nandakumar, et al., [14] trained a typical two-layered neural network using FHE which was implemented using the open-source library HElib introduced by Halevi & Shoup [15] for encryption.

An improved FHE scheme based on HElib was proposed by Sun, et al., [16], and a private hyper-plane decision-based classification and private Naïve Bayes Classification were implemented using the multiplicative homomorphic and additive homomorphic encryption. Using the proposed FHE scheme they implemented a private decision tree classification with the proposed FHE scheme.

Lee, et al [17] implemented a standard ResNet-20 model using the residue number system RNS-CKKS scheme, a fully homomorphic encryption scheme that is a variant of the CKKS scheme using the SEAL library 3.6.1.

A three-participant PPEML using additively homomorphic encryption consisting of data contributors, an outsourced server, and a data analyst was proposed by Kuri et al., [18] where the data contributor preprocesses and encrypts the data, the outsourced server computes the hidden layer outputs and the data analyst uses them to find the hidden connection weighs.

Wang, et al., [19] proposed a fully homomorphic Extreme Learning machine called Homo-ELM where the model is trained on the unencrypted data and the trained model is encrypted. To this trained model, encrypted data is provided and the encrypted model makes predictions. The idea of Homo-ELM is to apply in the cloud searching tasks.

A CKKS-based PPEML was proposed by Li & Huang [20]. The result of this research is that the users can encrypt the data and send it to service providers for analysis and prediction. The researchers also proposed a Single Instruction Multiple Data (SIMD) to reduce the user's waiting time by reducing the number of homomorphic operations.

Yu & Cao, [21] proposed chaotic Hopfield neural networks for encryption with time-varying delay. Binary Sequences are generated from the chaotic neural network which will be used for the masking of plain text. A chaotic color image encryption algorithm is proposed by Wang & Li [22] which uses a composite chaotic map combined with a staged Logistic Map and Tent Map.

## 3. Proposed Methodology

In this section, the types of homomorphic encryption and its concepts, along with the CKKS scheme proposed by Cheon, et al [23], which we employed for implementing the FHE, are explained. Later, we explain the original unencrypted ELM and describe our proposed Chaotic ELM and Privacy-Preserving Chaotic ELM.

### 3.1 Homomorphic Encryption

A special type of encryption scheme known as Homomorphic Encryption allows us to perform computation on the encrypted data. The encrypted data need not be decrypted at any point during computation, as mentioned by Acar, A., et al., [24] whereas there is a need for decryption in the other encryption schemes. Additive and Multiplicative homomorphism are supported by homomorphic encryption.

$$E(m_1 + m_2) = E(m_1) + E(m_2) \text{, and } E(m_1 * m_2) = E(m_1) * E(m_2)$$

where E is the encryption scheme and m1 and m2 are plain text. This implies that the homomorphically encrypted addition or multiplication of two numbers is equivalent to the addition or multiplication of two numbers that are individually homomorphically encrypted.

The homomorphic encryption scheme is classified into three categories based on the type, and the number of operations performed on the encrypted data:

### 3.1.1 Partially Homomorphic Encryption (PHE)

In this scheme, only similar kinds of operations, either addition or multiplication, can be performed any number of times on the encrypted data. RSA (multiplicative homomorphism) by Nisha & Farik [25], ElGamal (multiplicative homomorphism) Haraty, et al., [26], and Paillier (additive homomorphism) by Nassar, et al., [27] are some of the examples of PHE. Private Information Retrieval (PIR) and E-Voting make use of the PHE scheme.

### 3.1.2 Somewhat Homomorphic Encryption (SHE)

A limited number of both addition and multiplication operations on the encrypted data are allowed in this scheme. Some examples of SHE are Boneh-Goh-Nissim (BGN) and Polly Cracker Scheme.

### 3.1.3 Fully Homomorphic Encryption

An unlimited number of additions and multiplications can be performed on the encrypted data using the FHE scheme. But the drawback is it requires high end-resources and has computational complexity, as explained by Chialva & Dooms [28]. The concept of FHE, along with a general framework, was first introduced by Gentry [29] to obtain an FHE scheme. There are mainly three FHE families: Ideal lattice-based over integers van Dijk, et al., [30], Ring Learning With Errors (RLWE) by Brakerski & Vaikuntanathan [31], and NTRU-like López-Alt, et al., [32]. We implemented the Cheon-Kim-Kim-Song (CKKS) Scheme, whose security is based on the hardness assumption of the RLWE.

## 3.2 CKKS Scheme

Cheon-Kim-Kim-Song (CKKS) mainly works on an approximation of arithmetic numbers and is called a levelled homomorphic encryption scheme. The number of multiplications to be performed on the encrypted data is decided by the selection of parameters beforehand. That is why it is called levelled homomorphic encryption. This scheme works on the vector of real numbers and not on the scalar values. It is based on the library Homomorphic Encryption for Arithmetic of Approximate Numbers (HEAAN), which was first introduced by Cheon, et al [23]. The algorithms in HEAAN are implemented in C++, and it is an open-source homomorphic encryption library. In this paper, the CKKS scheme was used as we can encrypt the real numbers and perform the arithmetic operations and get approximate values to the original result.

### 3.2.1 Encryption in CKKS

The encryption process in CKKS Scheme happens in two steps. The vector of real numbers is encoded into a plain-text polynomial, and the plain text polynomial is encrypted into a ciphertext.

### 3.2.2 Decryption in CKKS:

The decryption also happens in two steps. The ciphertext is decoded into a plain-text polynomial in the first operation. The plain text polynomial is then decrypted to a vector of real numbers. The encryption and decryption process in the CKKS scheme is depicted in Figure 1.

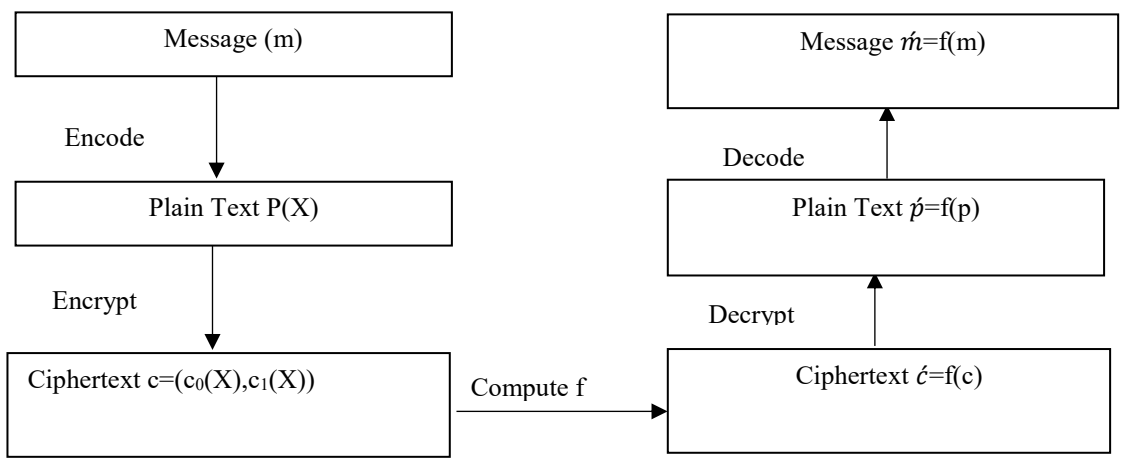

**Fig 1. Block Diagram of the Encryption and Decryption in CKKS Scheme.**

### 3.2.3 Parameters in CKKS:

The CKKS parameters decide the computational complexity and the privacy level of the model. These are as follows:

1 **Scaling Factor:** This defines the encoding precision for the binary representation of the number.
2 **Polynomial modulus degree:** This parameter is responsible for the number of coefficients in plain text polynomials, size of ciphertext, computational complexity, and security level. The degree should always be in the power of 2, for eg., 1024, 2048, 4096,…
   The higher the polynomial modulus degree, the higher the security level achieved. But, it will also increase the computational time.
3 **Coefficient Modulus sizes:** This parameter is a list of binary sizes. A list of binary sizes of those schemes will be generated which is called coefficient modulus size. The length of the list indicates the number of multiplications possible. The longer the list the lower the level of security of the scheme. The prime numbers in the coefficient modulus must be congruent to 1 modulo 2 * polynomial modulus degree.

### 3.2.4 Keys in CKKS

The scheme generates different types of keys which are handled by a single object called context. The keys are as follows:

1. **Secret Key:** This key is used for decryption and should not be shared with anyone.
2. **Public Encryption Key:** This key is used for the encryption of the data.
3. **Reliniearization Keys:** In general the size of the new ciphertext is 2. If there are two ciphertexts with sizes X and Y, then the multiplication of these two will result in the size getting as big as X + Y – 1. The increase in the size increases noise and also reduces the speed of multiplication. Therefore, Relinearization reduces the size of the ciphertexts back to 2 and this is done by different public keys which are created by the secret key owner.

### 3.3 Overview of the original unencrypted ELM

The ELM is a simple three-layered architecture consisting of the input layer, hidden layer, and output layer. The network consists of weight values between the input and hidden layer, and each neuron in the hidden layer has a bias. All the weight and bias values are generated randomly using uniform distribution in the range of 0 to 1, and Sigmoid is used as the activation function. The main task in the ELM model is to estimate the weights between the hidden and output layer using the Least Squares Method.

The algorithm to train ELM is as follows [33]:

1. Select the number of hidden and output nodes and initialize all the weights and biases randomly using uniform distribution in (0,1).
2. Biases will be added to the dot product of the input feature values with the weight values between the input and hidden layer. The sigmoid activation function is applied to the obtained result at the hidden layer.
3. The weights between the hidden and output layer are estimated using the Least Squares Method.
4. The mathematical representation of the ELM model is as follows:

$$\sum_{i=1}^{\widetilde{N}} \beta_i g_i(x_j) = \sum_{i=1}^{\widetilde{N}} \beta_i g_i(w_i \cdot x_j + b_i) = o_j$$

where j = 1,2,3,….., N, N is the total number of samples.

$w_i = [w_{i1}, w_{i2}, …..w_{in}]^T$ is the weight vector that connects the input nodes and $i^{th}$ hidden node.

$\beta_i = [\beta_{i1}, \beta_{i2}, ….\beta_{im}]^T$ is the weight vector that connects the $i^{th}$ hidden and the output nodes.

$b_i$ is the bias value of the $i^{th}$ hidden node.

$o_j$ is the output of the particular sample.

$\widetilde{N}$ is the number of hidden nodes.

g is the activation function.

5. The $\tilde{N}$ hidden nodes with the activation function g(x) can approximate the N samples with an error of zero which means that $\sum_{j=1}^{\tilde{N}} ||o_j - t_j|| = 0$,

So we can say that there exists $\beta_i$, $w_i$, and $b_i$ such that,

$$\sum_{i=1}^{\tilde{N}} \beta_i g_i(w_i \cdot x_j + b_i) = t_j$$

The above equation can be written as:

$$H\beta = T \text{ ---------------------------------- (1)}$$

Where $H(w_1,\ldots,w_{\tilde{N}}, b_1,\ldots b_{\tilde{N}}, x_1,\ldots x_N) = \begin{bmatrix} g(w_1 \cdot x_1 + b_1) & \ldots & g(w_{\tilde{N}} \cdot x_1 + b_N) \\ \vdots & \ldots & \vdots \\ g(w_1 \cdot x_N + b_1) & \ldots & g(w_{\tilde{N}} \cdot x_N + b_{\tilde{N}}) \end{bmatrix}$

$\beta = \begin{bmatrix} \beta_1^T \\ \vdots \\ \beta_{\tilde{N}}^T \end{bmatrix}$ and $T = \begin{bmatrix} t_1^T \\ \vdots \\ t_N^T \end{bmatrix}$

Here the matrix H is the output of the hidden layer in the neural network.

6. We need to calculate the $\beta$ in the equation $H\beta = T$

The smallest norm least squares solution of the above linear system is

$$\hat{\beta} = H^+ T \text{ ---------------------------------------(2)}$$

Where $H^+$ is the Moore-Penrose generalized inverse of matrix H.

The value of $H^+$ can be calculated by using the following formula:

$$H^+ = (H * H^T)^{-1} * H^T \text{ ------------------------- (3)}$$

7. By substituting $H^+$ in equation (2) we obtain the $\hat{\beta}$ which will be used to predict the outputs on the test data.

## 3.4 Proposed Chaotic Extreme Learning Machine and Privacy-Preserving Chaotic Extreme Learning Machine

### 3.4.1 Chaotic Extreme Learning Machine

In this paper, we proposed a Chaotic Extreme Learning Machine. Generally, in the Extreme Learning Machine, the weights and bias are randomly generated using a uniform distribution but in the proposed method we generate the weights and bias using the Logistic Map. The formula for the Logistic Map is given as follows:

$$x_{n+1} = r * x_n * (1 - x_n)$$

where we have taken the 'r' as 4 because the above equation exhibits chaotic behavior only when the value of r is greater than or equal to 3.56; $x_0$ is generated from a uniform distribution (0,1). If 'r' gets greater than 4 then the range of values generated will leave the interval [0,1] and will diverge.

We have varied the number of hidden nodes in the architecture based on the input nodes and calculated the accuracy in each case.

The model is trained similar to the traditional ELM where we obtain the $\hat{\beta}$ values by solving the equation (2)

The training and testing of the chaotic ELM are explained in Algorithm 1 and Algorithm 2 respectively.

```
Algorithm 1: Training the Chaotic ELM
Input:
Training Data:   x₁………..xN
Weights and biases generated using Logistic Map: w_{1,….},w_{Ñ} & b_{1},….b_{Ñ}
Output:
Weights between the hidden and output layer: β̂_{1}……β̂_{Ñ}
Function TrainChaoticELM (x₁……xN, w_{1,….},w_{Ñ} & b_{1},….b_{Ñ})
1. For each sample in the training set
2.      Perform dot product on the input feature values and
        each set of chaotic weight values.
3.      Add the bias to each resultant value
4.      Apply sigmoid on the final value and store all the
        values in the form of an NxÑ matrix and this is the
        H in equation 1.
5. Now calculate the β̂ values using equations (2) and (3)
6. end
```

```
Algorithm 2: Testing the Chaotic ELM
Input:
Test Data:   x'₁………..x'N
Weights and biases generated using Logistic Map: w_{1,….},w_{Ñ} & b_{1},….b_{Ñ}
Weights between hidden to output layer: β̂_{1}……β̂_{Ñ}
Output:
Predictions on the Test Data: y₁………yN
Function TestChaoticELM (x'₁…x'N, w_{1,….},w_{Ñ} , b_{1},….b_{Ñ}, β̂_{1}……β̂_{Ñ})
1. For each sample in the test set
2.      Perform dot product on the input feature values and
        each set of chaotic weight values.
3.      Add the bias to each resultant value
4.      Apply sigmoid on the final value and store all the
        values in the form of an NxÑ matrix and this is the
        in equation 1.
5. Multiply the H_{NxÑ} matrix with β̂_{1xÑ} to generate the output
   predictions
6. End
```

### 3.4.2 Privacy-Preserving Chaotic Extreme Learning Machine

We also proposed a Privacy-Preserving Chaotic Extreme Learning Machine and implemented FHE by using a library called TenSEAL (https://github.com/OpenMined/TenSEAL ). It provides a python API, but also maintains efficiency as most of its operations are implemented in C++. It performs encryption and decryption on the vector of real numbers using the CKKS scheme. It can perform various operations like addition, subtraction, multiplication, and dot product on encrypted vectors.

The weights and biases are generated as explained above in section 3.4.1. In this architecture, the weight between the input to hidden nodes $w_i$, and the biases $b_i$ are encrypted along with the input data.

The sigmoid activation function is used in the hidden layer of the ELM architecture but as we are working with the encrypted data we need to compute the polynomial approximation of the sigmoid activation function. The polynomial approximation is as follows [8]:

$$\sigma(x) = 0.5 + 0.197x - 0.004x^3 \quad \text{---------------- (4)}$$

The above equation approximates the sigmoid function in the range [-5,5].

The training of the ELM model is carried out on the encrypted data but to calculate the β values we need to decrypt the matrix H as we need to find the $H^+$ which is the Moore-Penrose generalized inverse of matrix H as explained in equation (3).

We need to find the inverse of H to calculate the $H^+$ but it is computationally very expensive to perform the inverse of the matrix H in the encrypted form. To reduce the complexity, following [18] [34] [35], where the authors have suggested and performed decryption, we also decrypted the matrix H before calculating $H^+$. Once we obtain the β values, we then encrypt these values along with the test data, and the final hence predictions are also in the encrypted form. Late,r these predictions are decrypted and the accuracy of the model is calculated.

The algorithm explains the parameters and the data that need to be encrypted in Algorithm 3 and the trainging and testing are explained in he Algorithm 4 and Algorithm 5.

```
Algorithm 3: Encryption
Input:
Unencrypted Training Data: $x_1$………..$x_N$
Unencrypted Test Data: $x'_1$………..$x'_N$
Unencrypted Weights and biases generated using Logistic Map: $w_1,…,w_{\widetilde{N}}$ & $b_1,….b_{\widetilde{N}}$
Output:
Encrypted Training Data:  $E(x_1)$………..$E(x_N)$
Encrypted Test Data:  $E(x'_1)$………..$E(x'_N)$
Encrypted Weights and biases generated using Logistic Map: $E(w_1),….,E(w_{\widetilde{N}})$ & $E(b_1),….E(b_{\widetilde{N}})$
   1. For each sample in the training set
   2.     Encrypt each sample as a CKKS Vector E(x)
   3. For each sample in the test set
   4.     Encrypt each sample as a CKKS Vector E(x')
   5. For each weight vector in the set of weights
   6.     Encrypt each weight vector as a CKKS Vector E(w)
   7. Encrypt the bias values as a CKKS vector E(b)
   8. End
```

```
Algorithm 4: Training Encrypted Chaotic ELM
Input:
Encrypted Training Data:  E(x₁)………..E(x_N)
Encrypted Weights and biases: E(w₁),….,E(w_Ñ) & E(b₁),….E(b_Ñ)
Output:
Encrypted weight values between the hidden and output layer:
E(β̂₁)……E(β̂_Ñ)
Function TrainEncryptedChaoticELM (x₁……x_N, w₁,….,w_Ñ & b₁,….b_Ñ)
  1. For each sample in the Encrypted training set
  2.     Perform dot product on the Encrypted input sample
         vector E(x_j) and the encrypted weight vectors
         E(w_i) and add the encrypted bias E(b to the
         resultant value.
  3.     Apply the approximate sigmoid function as in
         equation (4) on the final value and store all the
         values in the form of an NxÑ matrix and this is the
         H in equation 1.
  4. Decrypt the H matrix to calculate the  β̂ values as
     suggested in [18][34][35].
  5. Calculate the β̂ values using equations (2) and (3).
  6. Encrypt the obtained β̂ values as a CKKS vector E(β̂).
  7. End
```

```
Algorithm 5: Testing Encrypted Chaotic ELM
Input:
Encrypted Test Data:   E(x'_1)………..E(x'_N)
Encrypted Weights and biases: E(w_1),….,E(w_Ñ) & E(b_1),….E(b_Ñ)
Encrypted weight values between the hidden and output layer:
E(β̂_1)……… E(β̂_Ñ)
Output:
Encrypted Predictions on the Encrypted Test Data: E(y_1)………E(y_N)
Function TrainEncryptedChaoticELM (x_1……x_N, w_1,…,w_Ñ & b_1,….b_Ñ)
  1. For each sample in the Encrypted Test set
  2.     Perform dot product on the Encrypted input sample
         vector E(x_j) and the encrypted weight vectors
         E(w_i) and add the encrypted bias E(b to the
         resultant value.
  3.     Apply the approximate sigmoid function as in
         equation (4) on the final value and store all the
         values in the form of an NxÑ matrix and this is the
         H in equation 1.
  4. Multiply the encrypted E(H_{NxÑ}) matrix with E(β̂_{1xÑ}) to
     generate the Encrypted output predictions E(ŷ)
  5. End
```

## 4 Datasets Description

The features of all the datasets are presented in the Appendix.

### 4.1 Health Care Datasets

#### 4.1.1 Breast Cancer Coimbra Dataset

This dataset's features are anthropometric data and parameters generally gathered in a routine blood analysis. There are ten attributes, including the target variable, namely, the presence or absence of breast cancer, and 116 instances in this dataset [36]. Out of the 116 instances, 52 instances are the people who are healthy, and 64 instances are the people who are the risk of breast cancer. The description of the features is provided in Table A.1.

#### 4.1.2 Fertility Dataset

This dataset contains information about the semen samples provided by 100 volunteers. This dataset has 100 instances and ten features, including the target variable, namely, whether the diagnosis was Normal or Altered [37]. Out of the 100 instances, 88 instances are the samples whose output was Normal, and 12 instances are the samples whose output was Altered. The description of the features is provided in Table A.2.

### 4.1.3 Heart Disease Dataset

This dataset has 303 instances and 14 features that include the target variable, namely, whether the person has heart disease or not. Out of the 303 instances, 138 are the people who would not be affected by heart disease, and 165 are those who would be affected by heart disease. The description of the features is provided in Table A.3.

### 4.1.4 Diabetes Dataset

This dataset is mainly for the female gender and has 768 instances and nine features that include the target variable, namely, whether the patient has diabetes or not [38]. Out of the 768 instances, 500 instances are the people who are negative for diabetes, and 268 are the instances that are positive for diabetes. The description of the features is provided in Table A.4.

### 4.1.5 Haberman's Survival Dataset

The dataset contains samples collected from a study conducted between 1958 and 1970 at the University of Chicago's Billing Hospital on the patients who had survived after undergoing surgery for breast cancer. This dataset has 306 instances and four features, including the target variable: Survival Status of the Patient [39]. Of 306 cases, 225 are patients who survived more than five years, and 81 are people who died within five years. The description of features is provided in Table A.5.

## 4.2 Financial Datasets

### 4.2.1 BankNote Authentication Dataset

In this dataset, the data were extracted from images that were taken from genuine and forged banknotes. Wavelet transform was used to extract features from images. This dataset has 1372 instances and five features, including the target variable, whether the Note is authentic or not [40]. Out of 1372 instances, 762 instances are genuine Notes, and 610 instances are Notes which are forged. The description of the features is provided in Table A.6.

### 4.2.2 Qualitative Bankruptcy Dataset

In this dataset, there are 250 instances and seven features, including the target variable, namely, whether a bank is bankrupt or non-bankrupt [41]. Out of the 250 instances, 143 instances are non-bankrupt banks, and 107 are bankrupt. The description of the features is provided in Table A.7.

# 5 Results and discussion

A system with the configuration: HP Z8 workstation with Intel Xeon (R) Gold 6235R CPU processor, Ubuntu 20.04lts, and RAM of 376.6 GB was used to carry out all the experiments. Accuracy is taken as the performance metric.

We performed standardization on the features of the Breast Cancer Dataset, Fertility Dataset, Heart Disease Prediction Dataset, Diabetes Dataset, BankNote_Authentication Dataset, and Haberman's Survival Dataset. Standardization was not performed on the Bankruptcy Prediction dataset as it has categorical values for all the features. So the labels of all the features are converted into a numeric form. The fertility dataset is imbalanced with a distribution of Class '1': 12 and Class '0': 88. We applied the SMOTE [42] balancing technique and balanced the dataset with 88 samples in each class.

The predictions from the ELM model are continuous and to classify them into one of the two classes, we have used a threshold value of 0.5. All the prediction values less than 0.5 are classified as Class '0' and the values greater than 0.5 belong to Class '1'. There is also one more variation to it where we have applied the sigmoid function to the prediction values. Accuracy has been calculated on both the variations of the predictions.

A total of four variations of experiments have been carried out on the ELM: Encrypted Chaotic ELM, Unencrypted Chaotic ELM, Encrypted Traditional ELM, and Unencrypted Traditional ELM.

**5.1 Health Care Dataset Results**

In the health care datasets, the Chaotic version of the Encrypted and Unencrypted Extreme Learning Machine has performed slightly better or is similar to the Encrypted and Unencrypted version of the Traditional Extreme Learning Machine on most of the datasets.

In the Breast Cancer dataset, the Unencrypted version of both Chaotic and Traditional ELM yield similar results, and Encrypted Chaotic ELM performs better than Encrypted Traditional ELM. But the Encrypted Chaotic ELM is unable to perform as well as the Unencrypted Chaotic ELM as there is a difference of 8% in the accuracy. The results are provided in Table 1.

In the Fertility Dataset, the Unencrypted Chaotic ELM has the best accuracy among all the variations. The Encrypted Chaotic ELM, Encrypted Traditional ELM, and Unencrypted Traditional ELM have similar accuracy. The Encrypted Chaotic ELM almost performs similarly to the Unencrypted Chaotic ELM as there is only a difference of 3% in accuracy. The results are provided in Table 2.

In the Heart Disease Prediction Dataset, the Encrypted Chaotic ELM provides the best accuracy among all the variations. The Unencrypted version of both Chaotic and Traditional ELM has similar results and the Encrypted Traditional ELM performs better than both the Unencrypted versions but it is not as good as the Encrypted Chaotic ELM. The results are provided in Table 3.

In the Diabetes Dataset, the Encrypted Traditional ELM gives the best result among all the variations. The Encrypted Chaotic ELM also gives almost similar results to the Encrypted Traditional ELM with a difference of just 1% in accuracy. The Unencrypted versions of Chaotic and Traditional ELM give similar results but they are not as good as the Encrypted version. The results are provided in Table 4.

In Haberman's Survival Dataset, the Unencrypted Chaotic ELM provides the best results among all the variations. The remaining variations have similar accuracy and there is only a 1% difference in the accuracy of Unencrypted Chaotic ELM and the other variations. The results are provided in Table 5.

**5.2 Finance Dataset Results**

The results of the chaotic version of the Encrypted and Unencrypted Extreme Learning Machine have not been encouraging in the Financial Datasets as it has not been able to outperform the Traditional Extreme Learning Machine Learning model.

In the BankNote Authentication Dataset, the Unencrypted Traditional ELM gives the best results with an accuracy of 95%. The next best results are achieved by the Unencrypted and Encrypted Chaotic ELM with an accuracy of 85% and 81% respectively. The Traditional ELM has the lowest accuracy among all the variations with an accuracy of 80%.

We can observe that even though the Chaotic is not the best, the Encrypted Chaotic ELM is slightly better than the Encrypted Traditional ELM and there is only a difference of 4% between the Unencrypted and Encrypted Chaotic ELM. The results are provided in Table 6.

In the Bankruptcy Prediction Dataset, all the versions of the ELM model have the same accuracy of 58%.

On the whole, we can observe that in the health care datasets, the chaotic version of Encrypted as well as the unencrypted ELM has performed either better or similar to the Traditional version of the Encrypted and Unencrypted ELM but not even on one single health care dataset the accuracy of Chaotic ELM has been less than the Traditional ELM.

But in the case of BankNote Authentciation dataset which belongs to Finance, the Unencrypted Traditional ELM has performed the best but Chaotic Encrypted ELM performed better than the Encrypted Traditional ELM and on the Bankruptcy datasets, both Chaotic and Traditional ELM performed similarly.

The reason why Chaotic ELM performs better than Traditional ELM is that the random weight values between the input and hidden layer and bias for each hidden node are generated using a uniform distribution in the Traditional ELM. In the proposed Chaotic ELM the weight and bias values are generated using a logistic map that does not follow a uniform distribution.

The Encrypted ELM consumes a lot more time than the Unencrypted ELM in both Chaotic and Traditional ELM. But both Chaotic and Traditional ELM takes almost similar time in Unencrypted as well as Encrypted time. The time complexity increases especially in the Encrypted version as we increase the number of nodes in the Hidden Layer.

The coefficient modulus and polynomial modulus degree were taken as [40, 21, 21, 21, 21, 21, 21, 40] and 8192 respectively and the global scale was maintained as $2^{21}$. All the datasets were encrypted with the same set of parameters. A max bit count of 218 is provided by the polynomial modulus degree 8192. This means that sum of all the coefficient values should be less than or equal to 218. The number of possible multiplications supported by the scheme and the intermediate primes is responsible for the rescaling of the cipher text by the primes 1 to 7. The main aim of rescaling is to reduce the noise in the cipher text and keep the scale constant. The value of the global scale should be less than or equal to the intermediate primes and we have chosen an equal of 21 for both intermediate primes as well as the global scale. The first value of the coefficient modulus which is 40 in our case decides the size by which the plain text must be bounded and the last prime should be as large as the other primes.

## Table 1. Breast Cancer Dataset

| Hidden Nodes | Encrypted Chaotic ELM | | | Unencrypted Chaotic ELM | | | Encrypted Traditional ELM | | | Unencrypted Traditional ELM | | |
|---|---|---|---|---|---|---|---|---|---|---|---|---|
| | Linear | Sigmoid | Time | Linear | Sigmoid | Time | Linear | Sigmoid | Time | Linear | Sigmoid | Time |
| 2 | 0.54 | 0.54 | 27.72 | 0.375 | 0.54 | 0.28 | 0.54 | **0.54** | 27.74 | 0.50 | 0.54 | 0.22 |
| 3 | 0.45 | 0.54 | 41.83 | 0.50 | 0.58 | 0.20 | 0.54 | 0.54 | 42.47 | 0.50 | 0.54 | 0.20 |
| 4 | 0.29 | **0.58** | 61.74 | 0.58 | 0.54 | 0.20 | 0.50 | 0.50 | 62.57 | 0.45 | 0.54 | 0.20 |
| Same as input Nodes | 0.45 | 0.50 | 224.10 | 0.66 | **0.66** | 0.22 | 0.54 | 0.50 | 229.52 | **0.66** | 0.58 | 0.21 |

## Table 2. Fertility Dataset

| Hidden Nodes | Encrypted Chaotic ELM | | | Unencrypted Chaotic ELM | | | Encrypted Traditional ELM | | | Unencrypted Traditional ELM | | |
|---|---|---|---|---|---|---|---|---|---|---|---|---|
| | Linear | Sigmoid | Time | Linear | Sigmoid | Time | Linear | Sigmoid | Time | Linear | Sigmoid | Time |
| 2 | 0.66 | 0.50 | 41.93 | 0.50 | 0.50 | 0.20 | 0.55 | 0.50 | 43.59 | 0.44 | 0.50 | 0.20 |
| 3 | **0.69** | 0.50 | 64.85 | 0.50 | 0.50 | 0.21 | **0.69** | 0.50 | 67.45 | 0.47 | 0.50 | 0.22 |
| 4 | 0.58 | 0.50 | 92.42 | 0.38 | 0.50 | 0.20 | 0.63 | 0.50 | 97.32 | 0.50 | 0.47 | 0.30 |
| Same as input Nodes | 0.66 | 0.50 | 343.91 | **0.72** | 0.50 | 0.23 | 0.66 | 0.55 | 340.33 | **0.69** | 0.52 | 0.25 |

**Table 3. Heart Disease Prediction Dataset**

| Hidden Nodes | Encrypted Chaotic ELM | | | Unencrypted Chaotic ELM | | | Encrypted Traditional ELM | | | Unencrypted Traditional ELM | | |
|---|---|---|---|---|---|---|---|---|---|---|---|---|
| | Linear | Sigmoid | Time | Linear | Sigmoid | Time | Linear | Sigmoid | Time | Linear | Sigmoid | Time |
| 3 | 0.65 | 0.54 | 125.82 | 0.50 | 0.54 | 0.25 | 0.55 | 0.54 | 128.04 | 0.54 | 0.57 | 0.23 |
| 4 | 0.62 | 0.55 | 176.25 | 0.52 | 0.52 | 0.27 | 0.63 | 0.54 | 176.98 | 0.50 | 0.54 | 0.21 |
| 5 | **0.77** | 0.54 | 236.43 | 0.54 | 0.54 | 0.23 | **0.65** | 0.54 | 235.96 | 0.55 | 0.54 | 0.21 |
| 6 | 0.55 | 0.50 | 310.82 | 0.50 | 0.50 | 0.29 | 0.62 | 0.54 | 307.67 | 0.45 | 0.50 | 0.37 |
| Same as input Nodes | 0.63 | 0.54 | 1155.25 | **0.60** | 0.57 | 0.23 | 0.63 | 0.59 | 1142.84 | **0.60** | 0.55 | 0.22 |

**Table 4. Diabetes Dataset**

| Hidden Nodes | Encrypted Chaotic ELM | | | Unencrypted Chaotic ELM | | | Encrypted Traditional ELM | | | Unencrypted Traditional ELM | | |
|---|---|---|---|---|---|---|---|---|---|---|---|---|
| | Linear | Sigmoid | Time | Linear | Sigmoid | Time | Linear | Sigmoid | Time | Linear | Sigmoid | Time |
| 2 | 0.71 | 0.36 | 175.29 | **0.64** | 0.35 | 0.23 | 0.68 | 0.35 | 175.61 | **0.64** | 0.35 | 0.22 |
| 3 | **0.72** | 0.46 | 272.90 | 0.63 | 0.35 | 0.21 | 0.66 | 0.38 | 271.83 | 0.63 | 0.34 | 0.32 |
| Same as input Nodes | 0.69 | 0.37 | 1200.22 | 0.63 | 0.35 | 0.23 | **0.73** | 0.37 | 1194.58 | 0.63 | 0.35 | 0.21 |

### Table 5. Haberman's Survival Dataset

| Hidden Nodes | Encrypted Chaotic ELM | | | Unencrypted Chaotic ELM | | | Encrypted Traditional ELM | | | Unencrypted Traditional ELM | | |
|---|---|---|---|---|---|---|---|---|---|---|---|---|
| | Linear | Sigmoid | Time | Linear | Sigmoid | Time | Linear | Sigmoid | Time | Linear | Sigmoid | Time |
| 1 | 0.70 | 0.2580 | 24.93 | **0.75** | 0.25 | 0.20 | 0.72 | 0.25 | 26.78 | **0.74** | 0.25 | 0.21 |
| 2 | 0.25 | 0.2580 | 51.47 | 0.70 | 0.25 | 0.22 | 0.25 | 0.25 | 53.36 | 0.70 | 0.25 | 0.22 |
| Same as input Nodes | **0.74** | 0.7419 | 89.75 | 0.74 | 0.25 | 0.24 | **0.74** | 0.67 | 92.22 | 0.74 | 0.29 | 0.23 |

### Table 6. BankNote Authentication Dataset

| Hidden Nodes | Encrypted Chaotic ELM | | | Unencrypted Chaotic ELM | | | Encrypted Traditional ELM | | | Unencrypted Traditional ELM | | |
|---|---|---|---|---|---|---|---|---|---|---|---|---|
| | Linear | Sigmoid | Time | Linear | Sigmoid | Time | Linear | Sigmoid | Time | Linear | Sigmoid | Time |
| 1 | 0.26 | 0.44 | 134.49 | 0.81 | 0.44 | 0.23 | 0.16 | 0.44 | 130.39 | 0.77 | 0.44 | 0.21 |
| 2 | 0.78 | 0.54 | 254.06 | 0.76 | 0.44 | 0.24 | 0.46 | 0.43 | 250.31 | **0.95** | 0.44 | 0.24 |
| Same as input Nodes | **0.81** | 0.53 | 665.34 | **0.85** | 0.44 | 0.23 | **0.80** | 0.67 | 647.09 | 0.94 | 0.45 | 0.21 |

### Table 7. Qualitative Bankruptcy Prediction Dataset

| Hidden Nodes | Encrypted Chaotic ELM | | | Unencrypted Chaotic ELM | | | Encrypted Traditional ELM | | | Unencrypted Traditional ELM | | |
|---|---|---|---|---|---|---|---|---|---|---|---|---|
| | Linear | Sigmoid | Time | Linear | Sigmoid | Time | Linear | Sigmoid | Time | Linear | Sigmoid | Time |
| 2 | 0.50 | **0.58** | 50.29 | 0.58 | **0.58** | 0.33 | 0.58 | **0.58** | 51.67 | 0.58 | **0.58** | 0.240 |
| 3 | 0.48 | 0.58 | 81.31 | 0.58 | 0.58 | 0.26 | 0.48 | 0.58 | 83.67 | 0.56 | 0.58 | 0.612 |
| Same as input Nodes | 0.50 | 0.58 | 230.77 | 0.48 | 0.58 | 0.20 | 0.54 | 0.58 | 234.09 | 0.56 | 0.58 | 0.313 |

# 6 Conclusions

In a first-of-its-kind study, a Fully Homomorphic Encrypted Chaotic Extreme Learning Machine is proposed for the binary classification task. The model is highly protected as the input data along with other parameters such as weights and biases are encrypted. There is only a slight compromise on the security where we decrypt the output of the hidden layer to compute the inverse of the matrix to facilitate estimating the weights between the hidden and the output layer.

The time consumed for the ELM increases as we increase the number of hidden nodes and also it changes with a change in the number of samples and features in the dataset. From our experiments, we conclude that the Chaotic version of the Encrypted Extreme Learning Machine is not the best in all the scenarios but it has performed well on most of the datasets and is slightly better than the Traditional Encrypted Extreme Learning Machine.

# Appendix A

Table A.1 Breast Cancer Dataset

| Feature Description |
|---|
| 1. Age of the Patient |
| 2. Body Mass Index (kg/m$^2$) |
| 3. The glucose level in the body (mg/dL) |
| 4. Insulin level in the body (μU/mL) |
| 5. Homeostasis Model Assessment |
| 6. Leptin (ng/mL) |
| 7. Adiponectin (μg/mL) |
| 8. Resistin (ng/mL) |
| 9. Chemokine Monocyte Chemoattractant Protein 1 (MCP-1) |
| 10. Classification |

Table A.2 Fertility Dataset

| Feature Description |
|---|
| 1. Season in which the Analysis was performed |
| 2. Age at the time of analysis (18-36) |
| 3. Childish Disease (chicken pox, measles, mumps, polio) |
| 4. Accident or Serious Trauma |
| 5. Surgical Intervention |
| 6. High Fevers in the last year |
| 7. Frequency of Alcohol Consumption |
| 8. Smoking Habit |
| 9. Number of Hours spent sitting per day |
| 10. Output |

Table A.3 Heart Disease Dataset

| Feature Description |
|---|
| 1. Age of the Person |
| 2. Sex (Male or Female) |
| 3. Chest Pain Type |
| 4. Trestbps: resting blood pressure (in mm Hg on admission to the hospital) |
| 5. Cholesterol |
| 6. Fbs (fasting blood sugar > 120 mg/dl) |
| 7. restecg: resting electrocardiographic results |
| 8. thalach (maximum heart rate achieved) |
| 9. exang (exercise induced angina) |
| 10. Oldpeak (ST depression induced by exercise relative to rest) |
| 11. Slope (The slope of the peak ecercise ST segment) |
| 12. ca (number of major vessels covered by fluoroscopy) |
| 13. thal |
| 14. target (diagnosis of heart disease) |

Table A.4 Diabetes Dataset

| Feature Description |
|---|
| 1. Pregnancies (Number of times pregnant) |
| 2. Glucose (Oral Glucose Tolerance Test result) |
| 3. Blood Pressure (Diastolic Blood Pressure values in (mm Hg)) |
| 4. SkinThickness (Triceps skin fold thickness in (mm)) |
| 5. Insulin (2-Hour serum Insulin (μU/ml) |
| 6. BMI (Body Mass Index) |
| 7. Diabetes Pedigree Function |
| 8. Age (Age in Years) |
| 9. Class |

A.5 Haberman's Survival Datsaet

| Feature Description |
|---|
| 1. Age of patient at the time of Operation |
| 2. Patient's year of Operation |
| 3. Number of Positive Axillary Nodes Detected |
| 4. Survival Status |

Table A.6 BankNote Authentication Dataset

| Feature Description |
| --- |
| 1. The variance of Wavelet Transformed Image |
| 2. The skewness of Wavelet Transformed Image |
| 3. Curtosis of Wavelet Transformed Image |
| 4. Entropy of Image |
| 5. Class |

Table A.7 Bankruptcy Prediction

| Feature Description |
| --- |
| 1. Industrial Risk |
| 2. Management Risk |
| 3. Financial Flexibility |
| 4. Credibility |
| 5. Competitiveness |
| 6. Operating Risk |
| 7. Class |